\documentclass{llncs}
\usepackage {graphicx}

\begin{document}
\pagestyle{headings}
\pagestyle{empty}
\mainmatter

\title{Designing Digital Circuits for \\ the Knapsack Problem}

\author{Mihai Oltean\inst{1} 
\and Crina Gro\c san\inst{1}
\and Mihaela Oltean\inst{2}
}
\institute{
Department of Computer Science,\\
Faculty of Mathematics and Computer Science,\\
Babe\c s-Bolyai University, Kog\u alniceanu 1\\
Cluj-Napoca, 3400, Romania.\\
\email{\{moltean, cgrosan\}@cs.ubbcluj.ro}
\and 
David Prodan College,
Cugir, 2566, Romania.\\
\email{olteanmihaelaelena@yahoo.com}
}
\maketitle

\begin{abstract} Multi Expression Programming (MEP) is a Genetic 
Programming variant that uses linear chromosomes for solution encoding. A 
unique feature of MEP is its ability of encoding multiple solutions of a 
problem in a single chromosome. In this paper we use Multi Expression 
Programming for evolving digital circuits for a well-known NP-Complete 
problem: the knapsack (subset sum) problem. Numerical experiments show that 
Multi Expression Programming performs well on the considered test problems.

\end{abstract}

\section{Introduction}

The problem of evolving digital circuits has been deeply analyzed in the 
recent past \cite{miller1}. A considerable effort has been spent on evolving very 
efficient (regarding the number of gates) digital circuits. J. Miller, one 
of the pioneers in the field of the evolvable digital circuits, used a 
special technique called Cartesian Genetic Programming (CGP) \cite{miller1} for 
evolving digital circuits. CGP architecture consists of a network of gates 
(placed in a grid structure) and a set of wires connecting them. The results 
\cite{miller1} show that CGP is able to evolve digital circuits competitive to those 
designed by human experts.

In this paper, we use Multi Expression Programming (MEP)\footnote{ MEP 
source code is available from \url{https://mepx.org} or \url{https://mepx.github.io}.} \cite{oltean1} for evolving 
digital circuits. MEP is a Genetic Programming (GP) \cite{koza1} variant that uses 
linear chromosomes of fixed length. A unique feature of MEP is its ability 
of storing multiple solutions of a problem in a single chromosome. Note that 
this feature does not increase the complexity of the MEP decoding process 
when compared to other techniques that store a single solution in a 
chromosome. 

In this paper we present the way in which MEP may be efficiently applied for 
evolving digital circuits. We describe the way in which multiple digital 
circuits may be stored in a single MEP chromosome and the way in which the 
fitness of this chromosome may be computed by traversing the MEP chromosome 
only once. 

In this paper MEP is used for evolving digital circuits for a well-known 
NP-Complete \cite{garey1} problem: the knapsack (subset sum) problem. Since this 
problem is NP-Complete we cannot realistically expect to find a 
polynomial-time algorithm for it. Instead, we have to speed-up the existing 
techniques in order to reduce the time needed to obtain a solution. A 
possibility for speeding-up the algorithms for this problem is to implement 
them in assembly language. This could lead sometimes to improvements of over 
two orders of magnitude. Another possibility is to design and build a 
special hardware dedicated to that problem. This approach could lead to 
significant improvements of the running time. Due to this reason we have 
chosen to design, by the means of evolution, digital circuits for several 
instances of the knapsack problem. 

The knapsack problem may also be used as benchmarking problem for the 
evolutionary techniques which design electronic circuits. The main advantage 
of the knapsack problem is its scalability: increasing the number of inputs 
leads to more and more complicated circuits. The results show that MEP 
performs very well for all the considered test problems.

The paper is organized as follows. In section 2, the problem of designing 
digital circuits and the knapsack problem are shortly described. The Multi 
Expression Programming technique is presented in section 3. Several 
numerical experiments are performed in section 4.

\section{Problem Statement and Representation}

The problem that we are trying to solve in this paper may be briefly stated 
as follows:

\begin{center}
\textit{Find a digital circuit that implements a function given by its truth table.}
\end{center}

The gates that are usually used in the design of digital circuits along with 
their description are given in Table \ref{tab1}.

\begin{table}[htbp]
\caption{Function set (gates) used in numerical experiments. These 
functions are taken from \cite{miller1}}
\label{tab1}
\begin{center}
\begin{tabular}
{p{32pt}p{58pt}p{58pt}p{58pt}}
\hline
{\#}& 
Function& 
{\#}& 
Function \\
\hline
0& 
$a \cdot b$& 
5& 
$a \oplus \bar {b}$ \\
1& 
$a \cdot \bar {b}$& 
6& 
$a + b$ \\
2& 
$\bar {a} \cdot b$& 
7& 
$a + \bar {b}$ \\
3& 
$\bar {a} \cdot \bar {b}$& 
8& 
$\bar {a} + b$ \\
4& 
$a \oplus b$& 
9& 
$\bar {a} + \bar {b}$ \\
\hline
\end{tabular}
\end{center}
\end{table}

The knapsack problem (or the subset sum problem) may be stated as follows:

\textit{Let M be a set of numbers and a target sum k. Is there a subset S $ \subseteq $ M which has the sum k}?

The knapsack problem is a well-known NP-Complete problem \cite{garey1}. No 
polynomial-time algorithm is known for this problem.

Instead of designing a heuristic for this problem we will try to evolve 
digital circuits which will provide the answer for a given input.

In the experiments performed in this paper the set $M$ consists of several 
integer numbers from the set of consecutive integers starting with 1. For 
instance if the base set is {\{}1, 2, 3, 4, 5, 6, 7{\}} then $M$ may be {\{}2, 
5, 6{\}}. We will try to evolve a digital circuit that is able to provide the 
correct answer for all subsets $M$ of the base set.

The input for this problem is a sequence of bits. A value of 1 in position 
$k$ means that the integer number $k$ belongs to the set $M$, otherwise the number 
$k$ does not belong to the set $M$. 

For instance consider the consecutive integer numbers starting with 1 and 
ending with 7. The string 0100110 encodes the set $M$ = {\{}2, 5, 6{\}}. The 
numbers 1, 3, 4 and 7 do not belong to $M$ since the corresponding positions are 
0. The possible subsets of $M$ instance have the sum 2, 5, 6, 7, 8, 11 or 13. 
In our approach, the target sum is fixed and we are asking if is there a 
subset of given sum.

The number of training instances for this problem depends on the number of 
consecutive integers used as base for $M$. If we use numbers 1, 2 and 3, we 
have 2$^{3}$ = 8 training instances. If we use numbers 1, 2, 3, 4, 5, 6 and 
7, we have 2$^{7}$ = 128 training instances. In this case, whichever subset 
$M$ of {\{}1,\ldots ,7{\}} will be presented to the evolved circuit we have to 
obtain a binary answer whether the target sum $k$ may or not be obtained from a 
subset of $M$.

\section{Multi Expression Programming}

In this section, \textit{Multi Expression Programming} (MEP) \cite{oltean1} is briefly described. 

\subsection{MEP Representation}

MEP genes are represented by substrings of a variable length. The number of 
genes per chromosome is constant. This number defines the length of the 
chromosome. Each gene encodes a terminal or a function symbol. A gene 
encoding a function includes pointers towards the function arguments. 
Function arguments always have indices of lower values than the position of 
that function in the chromosome.

This representation is similar to the way in which \textbf{\textit{C}} and 
\textbf{\textit{Pascal}} compilers translate mathematical expressions into 
machine code \cite{aho1}.

The proposed representation ensures that no cycle arises while the 
chromosome is decoded (phenotypically transcripted). According to the 
proposed representation scheme the first symbol of the chromosome must be a 
terminal symbol. In this way only syntactically correct programs (MEP 
individuals) are obtained.\\

\textbf{Example}\\

A representation where the numbers on the left positions stand for gene 
labels is employed here. Labels do not belong to the chromosome, they are being 
provided only for explanation purposes.

For this example we use the set of functions $F$ = {\{}+, *{\}}, and the set of 
terminals $T$ = {\{}$a$, $b$, $c$, $d${\}}. An example of chromosome using the sets $F$ and 
$T$ is given below:

1: $a$

2: $b$

3: + 1, 2

4: $c$

5: $d$

6: + 4, 5

7: * 3, 6

\subsection{Decoding MEP Chromosomes and Fitness Assignment Process}

In this section it is described the way in which MEP individuals are 
translated into computer programs and the way in which the fitness of these 
programs is computed.

This translation is achieved by reading the chromosome top-down. A terminal 
symbol specifies a simple expression. A function symbol specifies a complex 
expression obtained by connecting the operands specified by the argument 
positions with the current function symbol.

For instance, genes 1, 2, 4 and 5 in the previous example encode simple 
expressions formed by a single terminal symbol. These expressions are:\\

$E_{1}=a$,

$E_{2}=b$,

$E_{4}=c$,

$E_{5}=d$,\\

Gene 3 indicates the operation + on the operands located at positions 1 and 
2 of the chromosome. Therefore gene 3 encodes the expression: \\

$E_{3}=a+b$. \\

Gene 6 indicates the operation + on the operands located at positions 4 and 
5. Therefore gene 6 encodes the expression:\\

$E_{6}=c+d$.\\

Gene 7 indicates the operation * on the operands located at position 3 and 
6. Therefore gene 7 encodes the expression:\\

$E_{7}$ = ($a+b)$ * ($c+d)$.\\

$E_{7}$ is the expression encoded by the whole chromosome.

There is neither practical nor theoretical evidence that one of these 
expressions is better than the others. Moreover, Wolpert and McReady \cite{wolpert1} 
proved that we cannot use the search algorithm's behavior so far for a 
particular test function to predict its future behavior on that function. 
This is why each MEP chromosome is allowed to encode a number of expressions 
equal to the chromosome length. Each of these expressions is considered as 
being a potential solution of the problem. 

The value of these expressions may be computed by reading the chromosome top 
down. Partial results are computed by dynamic programming and are stored in 
a conventional manner.

As MEP chromosome encodes more than one problem solution, it is interesting 
to see how the fitness is assigned.

Usually the chromosome fitness is defined as the fitness of the best 
expression encoded by that chromosome.

For instance, if we want to solve symbolic regression problems the fitness 
of each sub-expression $E_{i}$ may be computed using the formula:

\begin{equation}
\label{eq1}
f(E_i ) = \sum\limits_{k = 1}^n {\left| {o_{k,i} - w_k } \right|} ,
\end{equation}

\noindent
where $o_{k,i}$ is the obtained result by the expression $E_{i}$ for the 
fitness case $k$ and w$_{k}$ is the targeted result for the fitness case $k$. In 
this case the fitness needs to be minimized.

The fitness of an individual is set to be equal to the lowest fitness of the 
expressions encoded in chromosome:

\begin{equation}
\label{eq2}
f(C) = \mathop {\min }\limits_i f(E_i ).
\end{equation}

When we have to deal with other problems we compute the fitness of each 
sub-expression encoded in the MEP chromosome and the fitness of the entire 
individual is given by the fitness of the best expression encoded in that 
chromosome.

\subsection{Search Operators}

Search operators used within MEP algorithm are crossover and mutation. 
Considered search operators preserve the chromosome structure. All offspring 
are syntactically correct expressions. \\

\textbf{Crossover}\\

By crossover two parents are selected and are recombined. For instance, 
within the uniform recombination the offspring genes are taken randomly from 
one parent or another.\\

\textbf{Example}\\

Let us consider the two parents $C_{1}$ and $C_{2}$ given in Table \ref{tab2}. The two 
offspring $O_{1}$ and $O_{2}$ are obtained by uniform recombination as 
shown in Table \ref{tab2}.

\begin{table}[htbp]
\caption{MEP uniform recombination}
\label{tab2}
\begin{center}
\begin{tabular}
{p{55pt}p{49pt}p{49pt}p{49pt}}
\hline
\multicolumn{2}{p{104pt}}{\textit{Parents}} & 
\multicolumn{2}{p{99pt}}{\textit{Offspring}}  \\
\hline
$C_{1}$& 
$C_{2}$& 
$O_{1}$& 
$O_{2}$ \\
\hline
1: \textbf{\textit{b}} \par 2: \textbf{* 1, 1} \par 3: \textbf{+ 2, 1} \par 4: \textbf{\textit{a}} \par 5: \textbf{* 3, 2} \par 6: \textbf{\textit{a}} \par 7: \textbf{- 1, 4}& 
1: $a$ \par 2: $b$ \par 3: + 1, 2 \par 4: $c$ \par 5: $d$ \par 6: + 4, 5 \par 7: * 3, 6& 
1: $a$ \par 2: \textbf{* 1, 1} \par 3: \textbf{+ 2, 1} \par 4: $c$ \par 5: \textbf{* 3, 2} \par 6: + 4, 5 \par 7: \textbf{- 1, 4}& 
1: \textbf{\textit{b}} \par 2: $b$ \par 3: + 1, 2 \par 4: \textbf{\textit{a}} \par 5: $d$ \par 6: \textbf{\textit{a}} \par 7: * 3, 6 \\
\hline
\end{tabular}
\end{center}
\end{table}

\textbf{Mutation}\\

Each symbol (terminal, function of function pointer) in the chromosome may 
be target of mutation operator. By mutation some symbols in the chromosome 
are changed. To preserve the consistency of the chromosome its first gene 
must encode a terminal symbol.

\textbf{Example}

Consider the chromosome $C$ given in Table \ref{tab3}. If the boldfaced symbols are selected 
for mutation an offspring $O$ is obtained as shown in Table \ref{tab3}.

\begin{table}[htbp]
\caption{MEP mutation}
\label{tab3}
\begin{center}
\begin{tabular}
{p{81pt}p{67pt}}
\hline
$C$& 
$O$ \\
\hline
1: $a$ \par 2: * 1, 1 \par 3: \textbf{\textit{b}} \par 4: * 2, 2 \par 5: $b$ \par 6: +\textbf{ 3}, 5 \par 7: $a$& 
1: $a$ \par 2: * 1, 1 \par 3: \textbf{+ 1, 2} \par 4: * 2, 2 \par 5: $b$ \par 6: \textbf{+ 1}, 5 \par 7: $a$ \\
\hline
\end{tabular}
\end{center}
\end{table}

\subsection{MEP Algorithm}

In this paper we use a steady-state \cite{syswerda1} as underlying mechanism for Multi 
Expression Programming. The algorithm starts by creating a random population 
of individuals. The following steps are repeated until a stop condition is 
reached. Two parents are selected using a selection procedure. The parents 
are recombined in order to obtain two offspring. The offspring are 
considered for mutation. The best offspring replaces the worst individual in 
the current population if the offspring is better than the worst individual.

The algorithm returns as its answer the best expression evolved along a 
fixed number of generations.

\section{Numerical Experiments}

In this section several numerical experiments for evolving digital circuits 
for the knapsack problem are performed. The general parameters of the MEP 
algorithm are given in Table \ref{tab4}. Since different instances of the problem 
being solved will have different degrees of difficulty we will use different 
population sizes, number of genes in a chromosome and number of generations 
for each instance. Particular parameters are given in Table \ref{tab5}.

\begin{table}[htbp]
\caption{General parameters of the MEP algorithm for evolving digital circuits}
\label{tab4}
\begin{center}
\begin{tabular}
{p{131pt}p{150pt}}
\hline
\textbf{Parameter}& 
\textbf{Value} \\
\hline
Crossover probability& 
0.9 \\
Crossover type& 
Uniform \\
Mutations& 
5 / chromosome \\
Function set& 
Gates 0 to 9  (see Table \ref{tab1})\\
Terminal set& 
Problem inputs \\
Selection& 
Binary Tournament \\
\hline
\end{tabular}
\end{center}
\end{table}

\begin{table}[htbp]
\caption{Particular parameters of the MEP algorithm for different 
instances of the knapsack problem. In the second column the base set of 
numbers is given for each instance. In the third column the target sum is given.}
\label{tab5}
\begin{center}
\begin{tabular}
{p{16pt}p{40pt}p{25pt}p{55pt}p{61pt}p{50pt}p{60pt}}
\hline
{\#}& 
Set of numbers& 
Sum& 
Number of fitness cases& 
Population size& 
Number of genes& 
Number of generations \\
\hline
1& 
{\{}1\ldots 4{\}}& 
5& 
16& 
20& 
10& 
51 \\
2& 
{\{}1\ldots 5{\}}& 
7& 
32& 
100& 
30& 
101 \\
3& 
{\{}1\ldots 6{\}}& 
10& 
64& 
500& 
50& 
101 \\
4& 
{\{}1\ldots 7{\}}& 
14& 
128& 
1000& 
100& 
201 \\
\hline
\end{tabular}
\end{center}
\end{table}

Experimental results are given in Table \ref{tab6}. We are interested in computing 
the number of successful runs and the number of gates in the shortest 
evolved circuit.

\begin{table}[htbp]
\caption{Results obtained by MEP for the considered test problems. 
100 independent runs have been performed for all problems}
\label{tab6}
\begin{center}
\begin{tabular}
{p{18pt}p{67pt}p{45pt}p{100pt}p{100pt}}
\hline
{\#}& 
Set of numbers& 
Sum& 
Successful runs& 
Number of gates in  \par the shortest circuit \\
\hline
1& 
{\{}1\ldots 4{\}}& 
5& 
39 out of 100& 
3 \\
2& 
{\{}1\ldots 5{\}}& 
7& 
31 out of 100& 
5 \\
3& 
{\{}1\ldots 6{\}}& 
10& 
10 out of 100& 
11 \\
4& 
{\{}1\ldots 7{\}}& 
14& 
7 out of 100& 
21 \\
\hline
\end{tabular}
\end{center}
\end{table}

Table \ref{tab6} shows that MEP successfully found at least a solution 
for the considered test problems. The difficulty of evolving a digital 
circuit for this problem increases with the number of inputs of the problem. 
Only 20 individuals are required to obtain 39 solutions (out of 100 runs) 
for the instance with 4 inputs. In return, 1000 individuals (50 times more) 
are required to obtain 10 perfect solutions (out of 100 independent runs) 
for the instance with 7 inputs. Also the size of the evolved circuits 
increases with the number of problem inputs. However, due to the reduced 
number of runs we cannot be sure that we have obtained the optimal circuits. 
Additional experiments are required in this respect.

Due to the NP-Completeness of the problem it is expected that the number of 
gates in the shortest circuit to increase exponentially with the number of 
inputs.

\end{document}